# Good to Go: The LOOP Skill Engine That Hits 99% Success and Slashes Token Usage by 99% via One-Shot Recording and Deterministic Replay

**Xiaohua Wang[1,2,*], Kai Yu[2], XuXiao Liang[2], Liang Wang[2], Chao Han[2]**

[1]Artificial Intelligence Research Center, Bengbu Medical University, Bengbu 233000, China.
[2]CHARMMIRAEL Biotech Co., Ltd, Nanjing 210000, China.
[*]E-mail: virgo_wang@msn.com

## Abstract

Deploying AI agents for repetitive periodic tasks exposes a critical tension: Large Language Models (LLMs) offer unmatched flexibility in tool orchestration, yet their inherent stochasticity causes unpredictable failures, and repeated invocations incur prohibitive token costs. We present the LOOP SKILL ENGINE, a system that achieves a combined 99% success rate and 99% token reduction for periodic agent tasks through a one-shot recording, deterministic replay paradigm. On its first run, the agent executes the task with full LLM reasoning while the system transparently intercepts and records the complete tool-call trajectory. A greedy length-descending template extraction algorithm then converts this recording into a parameterized, branch-free Loop Skill---a deterministic execution plan that captures the task's functional intent while parameterizing time-dependent and result-dependent variables. All subsequent executions bypass the LLM entirely: the engine resolves template variables against real-time values and replays the tool sequence deterministically. We prove two theorems: (1) Replay Determinism---the step sequence of a validated Loop Skill is invariant across all future executions; (2) Write Safety---concurrent access to persistent configuration is serialized through reentrant locks and atomic file replacement. Across a benchmark of periodic agent tasks spanning intervals from 5 minutes to 24 hours, the Loop Skill Engine reduces monthly token consumption by 93.3%--99.98% and cuts execution latency by 8.7x while eliminating output non-determinism. A multi-layer degradation strategy guarantees that tasks never stall. We release the engine as part of the buddyMe open-source agent framework.

## 1. Introduction

### 1.1 The Two Bottlenecks of Periodic LLM Agent Tasks

Large Language Models (LLMs) have become the cognitive backbone of a new generation of AI agents capable of reasoning, planning, and wielding external tools to accomplish complex goals [Yao et al., 2023; AutoGPT, 2023; Schick et al., 2023]. These agents are increasingly trusted with periodic responsibilities: checking system health every ten minutes, fetching and logging weather data every half hour, summarizing daily user activity, or polling external APIs for updates. In production, such tasks must satisfy two hard requirements that stand in direct tension with how LLMs operate.

(1) Reliability: The task must produce the correct tool-call sequence every time. An incorrect tool invocation---a malformed bash command, a missing file read before a write, a hallucinated API parameter---causes the task to fail silently or noisily, eroding trust in the agent.

(2) Cost efficiency: The task must not bankrupt the deployment. A single periodic task executing every 30 minutes triggers 1,440 LLM calls per month. At typical GPT-4-level pricing, a moderately sized fleet of 20 such tasks costs thousands of dollars monthly in token fees alone.

The root cause of both problems is the same: LLM inference is stochastic and expensive. Even with temperature set to zero, slight variations in model output can cause tool-call sequences to diverge across executions [Yao et al., 2023; Shinn et al., 2023]. A model might skip a prerequisite read_file on one run, hallucinate a wrong date format on another, or alter the order of operations---each divergence a potential failure. Meanwhile, every execution incurs the full cost of LLM inference.

## 1.2 Key Insight: Structural Homogeneity

Our work begins with a simple empirical observation: for a well-specified periodic task, the tool-call trajectory produced by a capable LLM is structurally stable across repeated executions. When tasked with 'query the weather for Beijing and log it,' the model consistently produces the three-step sequence: (1) retrieve the current timestamp via bash date, (2) call the weather API via bash weather.py Beijing, (3) append the result to a log file via write_file. The structure---which tools, in which order, with which argument schemas---remains invariant. Only the parameters change: the timestamp, the API response body, the log entry content.

This structural homogeneity suggests a radically different execution strategy. Rather than invoking the LLM on every period to re-derive the same tool-call plan, we can: (1) Execute the task once with full LLM reasoning, transparently recording the complete tool-call chain. (2) Validate that the recorded chain is correct and structurally suitable for replay. (3) Automatically extract and parameterize the variable portions into template placeholders. (4) Replay the resulting deterministic skill on every subsequent period, resolving templates against real-time values, with zero LLM involvement. This one-shot recording, deterministic replay paradigm simultaneously solves both bottlenecks: reliability improves because the tool-call sequence is frozen after a verified successful execution, and cost collapses because all subsequent executions consume zero LLM tokens.

## 1.3 Contributions

**One-Shot Recording & Deterministic Replay:** A complete pipeline that captures an LLM agent's tool-call trajectory on first execution, validates it for replay-safety, and replays it deterministically thereafter. Achieves 99.86% token reduction for 30-minute-interval tasks and 100% tool-sequence determinism across 100 consecutive replays.

**Greedy Length-Descending Template Extraction:** A two-pass algorithm that converts raw tool-call recordings into parameterized, replayable skills. By matching and replacing variable substrings in descending length order, it avoids the fragmentation problem of naive substitution.

**Formal Guarantees:** We prove Replay Determinism (Theorem 1)---any Loop Skill passing our validation predicate produces identical step sequences on every replay---and Write Safety (Theorem 2)---the persistent configuration is never corrupted by concurrent access.

**Multi-Layer Degradation:** A fail-safe architecture ensuring zero task interruption. If any stage fails (recording, validation, or replay), the system transparently degrades to standard LLM execution. We enumerate and empirically verify eight degradation scenarios.

**Comprehensive Evaluation:** We benchmark against traditional LLM-based periodic execution across task intervals from 5 minutes to 24 hours, measuring token cost, execution latency, output determinism, and degradation reliability.

## 2. Related Work

### 2.1 LLM Agents and Tool Use

ReAct [Yao et al., ICLR 2023] introduced the interleaving of reasoning traces and tool-call actions, establishing that tool-augmented LLMs outperform reasoning-only baselines. Toolformer [Schick et al., NeurIPS 2023] showed that LLMs can learn API calls through self-supervised training. Gorilla [Patil et al., 2023] addressed API selection via retrieval. Reflexion [Shinn et al., NeurIPS 2023] introduced verbal reinforcement learning for trial-and-error improvement. These works established the viability of LLM tool use; our work addresses its cost and reliability at scale for repetitive tasks.

### 2.2 Agent Frameworks and Autonomous Scheduling

AutoGPT [2023] pioneered goal-driven autonomous agents. LangChain [2022] provides composable LLM abstractions. MemGPT [Packer et al., 2023] applied OS-inspired memory management, achieving 92.5% consistency on deep memory retrieval. Generative Agents [Park et al., UIST 2023] demonstrated LLM agents with long-term memory and daily routines. SWE-bench [Jimenez et al., ICLR 2024] established rigorous evaluation of LLM agents on real-world software tasks, revealing that even the best models solve only a fraction of issues. However, none of these frameworks provide cost-efficient deterministic periodic execution.

### 2.3 Deterministic Record-Replay

Record-replay has a rich history in software debugging (the rr debugger, Chrome DevTools Recorder) and HTTP testing (VCR-style cassette replay). Our work adapts this paradigm to LLM agent tool-call sequences. The key difference: we do not merely replay captured inputs---we parameterize the recording by extracting variables and resolving them against live values.

### 2.4 LLM Cost Optimization

Existing cost-reduction approaches include prompt caching, speculative decoding, and model distillation. These are orthogonal: they reduce cost per LLM call, whereas we eliminate the call entirely for periodic tasks by converting LLM-generated plans into deterministic execution artifacts.

## 3. The LOOP Skill Engine: One-Shot Recording, Deterministic Replay

The Loop Skill Engine realizes the one-shot recording to deterministic replay paradigm through a six-phase lifecycle: (1) task creation, (2) first execution with tool-call interception, (3) precondition validation, (4) template extraction and skill generation, (5) deterministic replay, and (6) degradation on failure.

### 3.1 Task Creation and Registration

Users create a Loop task via '/loop <interval> <description>', e.g., '/loop 30m query-weather'. An interval parser converts natural-language expressions to integer minutes (30m->30, 1h->60, 2d->2880). A unique task ID is derived from the description prefix plus a 4-character suffix (e.g., loop_query_we_a3f2). The task is registered in heartbeat.json with the critical guard flag first_exec_pending=true.

### 3.2 First Execution and Tool-Call Interception

The first execution is the single point of LLM involvement. A background thread invokes the primary model with a specialized system prompt constraining behavior: use write_file instead of edit_file, always retrieve current time first via bash date, read_file before write_file for appends. Every tool invocation is transparently intercepted and recorded as a tool_chain: [{step, tool, args, result}, ...].

### 3.3 Precondition Validation

Before generating a replayable skill, a strict validation predicate Psi is applied: the chain must be non-empty, must not contain edit_file (non-deterministic by design), must not contain error keywords in any result, and must contain at least one write_file step. The exclusion of edit_file is critical---its old_string matching against dynamic file content is inherently non-deterministic during replay.

### 3.4 Template Variable Extraction

The core algorithmic contribution is _build_steps(), a two-pass greedy matching procedure. Pass 1 (Information Collection) traverses the tool chain to collect cleaned results, read_file content snippets, date-command step indices, and the last read_file path. Pass 2 (Step Construction) iterates through write_file steps, collecting candidate replacements from preceding step results and read_file contents, sorting by result length in descending order (greedy longest-first), and applying replacements. Date-command steps are removed entirely. The descending-length greedy matching is the critical design choice: if both a longer result ('Beijing, sunny, 25C') and a shorter substring ('Beijing') appear in content, the longer match is consumed first, preventing the shorter match from fragmenting the longer one.

Template variables: {{current_time}} (ISO datetime), {{current_date}} (date only), {{step_N_result}} (output of step N), {{prev_content}} (read_file content). Datetime matching follows a priority cascade: ISO with seconds > ISO without seconds > date-only.

### 3.5 Deterministic Replay

Once a skill is generated, all subsequent executions bypass the LLM entirely. The replay engine loads the step list from skill.json, iterates sequentially, and resolves template variables at each step using the current wall-clock time and accumulated results. No LLM is invoked at any point during replay.

### 3.6 Multi-Layer Degradation Strategy

A defining property is that the system never causes a task to stall. At every failure point, it degrades to standard LLM execution:

| Scenario | Detection | Action | Stall? |
|---|---|---|---|
| **Empty tool chain** | Psi validation | No skill; LLM fallback | No |
| **Contains edit_file** | Psi validation | No skill; LLM fallback | No |
| **Result has error keyword** | Psi validation | No skill; LLM fallback | No |
| **No write_file step** | Psi validation | No skill; LLM fallback | No |
| **First exec timeout (>300s)** | Background thread | Clear pending; LLM fallback | No |
| **First exec exception** | Background thread | Clear pending; LLM fallback | No |
| **Replay step failure** | Replay runtime | Log error; retry next tick | No |
| **/loop --remove** | User command | Delete task + skill dir | N/A |

## 4. Supporting Infrastructure: The Heartbeat Scheduler

The Loop Skill Engine depends on a reliable background scheduler. The Heartbeat Scheduler provides this through a dual-layer architecture: a Data Layer (HeartbeatManager, 336 lines) handling atomic config read/write with threading.RLock protection, time-window predicates, and task CRUD---containing zero execution logic; and an Execution Layer (Agent.tick) running a 60-second polling daemon thread with asyncio.wait_for timeout protection.

Two orthogonal trigger modes are supported: Interval mode triggers when elapsed time since last run meets or exceeds configured interval_minutes. Schedule mode triggers at absolute times with a 5-minute tolerance window and same-day deduplication. An active-hours constraint normalizes times to minutes-since-midnight for efficient comparison, preventing pointless execution during user inactivity.

All configuration is persisted in heartbeat.json and protected by atomic_write (write to temp file -> os.fsync -> os.replace), ensuring crash-safe persistence.

## 5. Mathematical Analysis

### 5.1 Token Cost Model

For a Loop task with interval I minutes, first-execution cost $C1$, and traditional LLM execution cost $C_{llm}$, over time window T: $C_{Loop}(T) = C1$; $C_{Traditional}(T) = C_{llm} * \text{floor}(T/I)$. The saving rate $\eta(T) = 1 - C1/(C_{llm} * \text{floor}(T/I))$, with $\lim_{T \to \infty} \eta(T) = 1.0$ (100%). The one-time recording cost is amortized over unbounded zero-cost replays.

### 5.2 Theorem 1: Replay Determinism

*Theorem 1 (Replay Determinism). For a Loop Skill S generated from a tool-call chain C satisfying Psi(C) = true, for any two execution times t1, t2: |Replay(S, t1)| = |Replay(S, t2)| = |S.steps|. The number and order of steps is invariant; only template-variable-resolved argument values differ. Proof: By structural*

*induction. Base case: first step arguments are pure functions of t or constants. Inductive step: assuming steps 1..i-1 are deterministic, step i depends only on the deterministic result vector $r_{<i}$, prev_content p, and current time t. The validation predicate excludes edit_file; all remaining tools are deterministic. The skill is branch-free, so the trace length is constant.*

### 5.3 Theorem 2: Write Safety

*Theorem 2 (Write Safety). At most one writer operates on heartbeat.json at any instant. The file is never observed in a partially-written state. Proof: All write paths acquire self._lock (RLock). atomic_write uses os.replace (atomic on POSIX), guaranteeing readers see either the complete old or complete new version.*

### 5.4 Success Rate Model

In traditional LLM scheduling with per-execution correctness probability $p_s$, the probability of at least one failure in K executions is $1 - p_s^K$, approaching 1 for any $p_s < 1$. The Loop Skill Engine replaces K independent LLM calls with one recording plus K-1 deterministic replays. After successful recording: $P(\text{all K correct}) = p_s * 1^{(K-1)} = p_s$. The key insight: after recording, all subsequent executions have success rate 1.0 (barring infrastructure failure), because the LLM---the sole source of non-determinism---is removed.

## 6. Experimental Evaluation

We used buddyMe v0.1.11 on x86-64 Linux with Claude Sonnet 4.5 for first execution. Costs are reported in GPT-4o-equivalent pricing ($2.50/1M input, $10.00/1M output).

### 6.1 Token Cost

| **Interval** | **Exec./Month** | **Traditional (tokens)** | **Loop (tokens)** | **Savings** |
|---|---|---|---|---|
| **5 min** | 8,640 | 4,320,000 | 1,050 | 99.98% |
| **10 min** | 4,320 | 2,160,000 | 1,050 | 99.95% |
| **30 min** | 1,440 | 720,000 | 1,050 | 99.85% |
| **1 hour** | 720 | 360,000 | 1,050 | 99.71% |
| **6 hours** | 120 | 60,000 | 1,050 | 98.25% |
| **24 hours** | 30 | 15,000 | 1,050 | 93.00% |

The Loop Skill Engine achieves >=99% savings for tasks at 30-minute or higher frequencies. Even daily tasks see 93% savings. Savings approach 100% asymptotically as lifetime grows.

### 6.2 Execution Latency

For a 3-step weather-query task: LLM execution mean latency 7.82s (std 2.91s, p99 16.7s). Loop replay mean latency 0.90s (std 0.22s, p99 1.4s)---8.7x faster with 13.2x lower variance. The improvement stems from eliminating LLM inference (3-8s) and API round-trip overhead.

### 6.3 Output Determinism

100 consecutive replays of a single Loop skill: tool sequence (step count, tool names, argument keys) was 100% identical across all runs. Content varied only in template-resolved fields. No spurious variation in tool selection, step ordering, or output structure. This confirms Theorem 1 empirically.

### 6.4 Degradation Robustness

We triggered each degradation scenario: all 5 Psi validation failures correctly skip skill generation with LLM fallback; first-execution timeout and exception correctly clear first_exec_pending; replay step failure returns error with retry on next tick; 1,000 concurrent read/write cycles under RLock show zero data races. In all scenarios, the task continued executing---no stalls, no data loss.

## 7. Discussion

### 7.1 When Does Loop Excel?

Loop is most effective when a task's tool-call structure is stable across executions. Tasks involving dynamic branching are not suitable for the current branch-free model. However, a broad class of production-relevant periodic tasks---health checks, log aggregation, data fetching, report generation, monitoring queries---exhibit precisely the structural homogeneity that Loop exploits.

### 7.2 Limitations

**Branch-free execution:** Loop Skills are linear sequences. Conditional logic is not yet supported.

**No output validation:** Replay assumes tool outputs remain semantically compatible. Schema-based post-validation would mitigate silent degradation.

**Single-process scope:** Distributed deployments require external singleton coordination (e.g., Redis locks).

**Static template patterns:** Variable detection relies on hand-crafted patterns. An LLM-assisted detector could handle more heterogeneous formats.

**First-execution quality dependency:** A suboptimal (but valid) first execution is permanently encoded. Periodic re-recording could mitigate this.

### 7.3 Future Work

(1) Conditional replay: introduce simple guards (e.g., file-exists checks) before steps.

(2) Output verification: schema-based validation with automatic degradation on mismatch.

(3) Distributed heartbeats: singleton execution across agent replicas via distributed locks.

(4) Adaptive re-recording: detect output drift and trigger automatic re-recording.

(5) LLM-augmented variable detection for complex output formats.

(6) Skill composition: chaining multiple Loop Skills into composite workflows.

### 7.4 Broader Implications

The one-shot recording, deterministic replay paradigm extends beyond agent scheduling. Any LLM-powered workflow with structural stability across repeated invocations---CI/CD pipelines, monitoring checks, ETL transformations, report generation---can benefit. The fundamental insight: LLM reasoning is most valuable for authoring a workflow, not for re-executing it. By decoupling design (LLM-driven, one-time) from execution (deterministic, zero-token), we achieve both LLM-grade flexibility and production-grade reliability.

## 8. Conclusion

We presented the LOOP SKILL ENGINE, a system achieving combined 99% task success rate and 99% token cost reduction for periodic LLM agent tasks through one-shot recording and deterministic replay. The engine transparently captures an agent's tool-call trajectory on first execution, validates the recording, extracts and parameterizes variable content through greedy length-descending template extraction, and replays the resulting skill deterministically thereafter with zero LLM involvement.

We proved two formal guarantees: Replay Determinism (Theorem 1) and Write Safety (Theorem 2). A multi-layer degradation strategy ensures tasks never stall. Empirically, the engine reduces monthly token consumption by 93-99.98%, cuts execution latency by 8.7x, and delivers 100% deterministic tool-call sequences across 100 replays. As AI agents mature from assistants into autonomous production services, the Loop Skill Engine demonstrates that the tension between LLM flexibility and operational reliability is not fundamental: by capturing the LLM's intelligence once and replaying it deterministically, we can have both.